\documentclass[sigconf]{acmart}

\AtBeginDocument{%
  }

\setcopyright{acmlicensed}
\copyrightyear{2018}
\acmYear{2018}
\acmDOI{XXXXXXX.XXXXXXX}

\acmConference[ACM ICAIF 2024]{From Prototype to Production:
Deploying Real-World AI / ML Models in the Financial Industry}{June 03--05,
  2018}{Woodstock, NY}
\acmISBN{978-1-4503-XXXX-X/18/06}

\begin{document}

\title{Hierarchical Fallback Architecture for High Risk Online Machine Learning Inference}

\author{Gustavo Polleti}
\email{gustavo.polleti@trustly.com}
\orcid{1234-5678-9012}
\affiliation{%
  \institution{Trustly}
  \city{S\~ao Paulo}
  \state{S\~ao Paulo}
  \country{Brazil}
}

\author{Marlesson Santana}
\email{marlesson.santana@trustly.com}
\orcid{1234-5678-9012}
\affiliation{%
  \institution{Trustly}
  \city{Ilhéus}
  \state{Bahia}
  \country{Brazil}
}

\author{Felipe Sassi Del Sant}
\email{felipe.sant@trustly.com}
\orcid{1234-5678-9012}
\affiliation{%
  \institution{Trustly}
  \city{Florianópolis}
  \state{Santa Catarina}
  \country{Brazil}
}

\author{Eduardo Fontes}
\email{eduardo.fontes@trustly.com}
\orcid{1234-5678-9012}
\affiliation{%
  \institution{Trustly}
  \city{Concord}
  \state{North Carolina}
  \country{USA}
}



\begin{abstract}
  Open Banking powered machine learning applications require novel robustness approaches to deal with challenging stress and failure scenarios. In this paper we propose an hierarchical fallback architecture for improving robustness in high risk machine learning applications with a focus in the financial domain. We define generic failure scenarios often found in online inference that depend on external data providers and we describe in detail how to apply the hierarchical fallback architecture to address them. Finally, we offer a real world example of its applicability in the industry for near-real time transactional fraud risk evaluation using Open Banking data and under extreme stress scenarios.
\end{abstract}



\received{20 February 2007}
\received[revised]{12 March 2009}
\received[accepted]{5 June 2009}

\maketitle

\section{Introduction}

Recent developments in AI/ML and the broadening of the applications across many industries has substantially transformed and optimized business impacts drastically. While the positive stories are more commonly shared, many aspects of technical deficiencies have also surfaced, given most organizations core systems were not originally planned to make real time decisions. These systems can fail unexpectedly in a variety of different ways. Notably, applications that rely on online inference are subject to their inability to keep up with the expected operating procedures while, now additionally, having to make tedious computational tasks for these AI/ML applications, typically resulting in timeouts, infrastructure outages and, often, failures in external dependencies such as third party data providers (external API calls)~\cite{sculley2015hidden}. When the underlying machine learning applications are presented with strong robustness requirements, fallback or fall-over strategies are needed to keep operations running, even in the event of unexpected failures. In finance, specifically applications that require real time risk mitigation decision making, having systems shortgages are subject to substantial financial losses, making the robustness of these systems required. If an ML system fails and a risky loan or a fraudulent transaction gets mistakenly approved, the financial institution (FI) can be severely harmed. The challenge increases if we consider that many modern ML systems used in financial decision making currently depend on Open Banking data~\cite{LongT0Z20}. Financial institutions work as external data providers in the context of Open Banking by providing relevant information for plenty of use cases, including risk evaluation. To illustrate, If a financial institution fails to provide the Open Banking data required to assess whether a transaction is fraudulent or not, the risk assessment should contain fallback strategies to keep the evaluation robust enough to prevent fraudsters from benefiting from potential outages. One possible approach is to develop ML models taking robustness into account so that it can produce a relevant inference even when part of its features are absent or present inaccuracies~\cite{MYLLYAHO2022111096}. However, robustness is often hard to guarantee in the model development stage~\cite{Einziger19} and the ML model itself can be unavailable due to an internal outage. In situations where the ML model unavailability persists, another approach is to route the request to a fallback model. The fallback model represents a variant of the main model that is developed taking into account a failure scenario. For example, if an external data provider call fails, it is expected that a given group of features will be affected. The fallback model could be trained without the affected feature group and so is not impacted  by the outage. Additionally, the fallback model represents a proxy to an infrastructure redundancy that can be useful if the main one faces downtime.

In this work, we consider the following failure scenarios: When some data required for the ML model is (1) absent due to unavailability in the data provider or (2) corrupted due to other types of failures, (3) the model inference itself is unavailable or (4) presents latency spikes. We propose a hierarchical fallback architecture to deal with all considered failure scenarios while meeting strict response time constraints. Our architecture distinguishes a single main model, which fulfills the requests in a non-failure scenario, and multiple fallback models that are activated in failure scenarios. We describe how to develop fallback models and organize them, coupled with retry policies, in the online inference setting. We present a case study that discusses the effectiveness of our hierarchical fallback architecture in a real risk evaluation use case using Open Banking data.

The paper is organized as follows: Section 2 presents the literature on robustness and discusses alternatives. Section 3 describes the hierarchical fallback architecture and the development of fallback models. Section 4 discusses the challenges and limitations of our approach considering Trustly’s Open Banking Payment real use case under stress scenarios, and offers concluding remarks.

\section{Background}

It is possible to find in the literature a diverse setting of methods on robustness in machine learning. Previous works focused on how to embed during model development robustness properties~\cite{Einziger19}. Some approaches employ adversarial techniques to produce models robust to corrupted data~\cite{Narodytska18}. Overall, it is hard to ensure that the model is robust enough. As we mentioned, ML can fail in unexpected ways and it is hard to anticipate all of them. Robustness validation and guarantees for ML remains an open problem, so even if a robustness technique is applied, it cannot ensure the model will behave properly in all failure scenarios. Furthermore, robustness properties at model development can only deal with failures in the upstream data sources, our failure types (1) and (2), and do not address failures in the model infrastructure, our failure types (3) and (4). Despite relevant results on making ML more reliable, robustness in ML has mostly focused on testing ML models, and less on the whole systems that utilize ML~\cite{zhang2020machine}. Recent studies with professionals from different industries suggest redundant and fallback models as an emerging architecture for robust ML systems~\cite{MYLLYAHO2022111096}. Redundant fallback models are useful when the main model faces downtime. We also argue that fallback models are more predictable than robustness techniques applied to the main model at development time. For example, in case a feature group is affected by a failure, a fallback model, that does not include features from the affected group, will not have its predictive performance affected. On the other hand, one cannot anticipate the performance drop of a model that uses the affected feature group, despite having some robustness properties, will have.

\section{Method}

\begin{figure*}
  \includegraphics[width=0.7\textwidth]{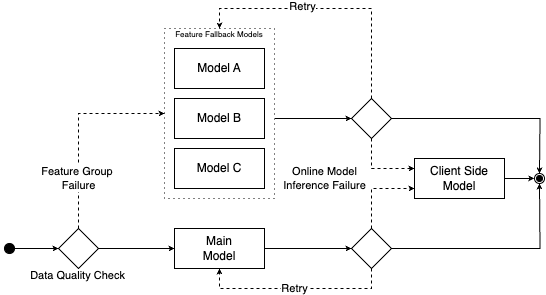}
  \caption{Hierarchical Fallback Architecture.}
  \label{fig:fallback_arch}
\end{figure*}

Figure \ref{fig:fallback_arch} presents the proposed architecture, composed of a main model and three fallback steps responsible for dealing with different failure scenarios. The first stage aims to deal with input data absent due to unavailability in the data provider, which can lead to the unavailability of one or more features groups. To handle this type of failure, we fallback to a similar model that was trained without the feature group. Basically, for each critical features group, we have a model version available to be used as a fallback in case of failure, where the weights of the features are recalibrated according to their availability for each model, supporting the maximization of signal extraction. This approach ensures that in the fallback event due to data provider failure, we will be using the best available model version to make the decision.

The second fallback step applies redundancy optimization using  any of the available models, the retry policy aims to mitigate problems of momentary service unavailability or latency spikes. The decision to retry is made at transaction time and takes into account transaction information, probability of a subsequent failure and also additional time in risk analysis. This approach ensures that, in the event that the model inference itself is not available due to some momentary instability, the decision to retry the main model before fallback takes into account other risk factors.

The last fallback step is the client side fallback model, where, only when all other steps are unavailable, this step is used. In our current architecture, all machine learning models are used from the same main software. In this case we implemented a risk model embedded in the main software, what we call Client Side Fallback Model. This model uses fewer features resulting in lesser memory to be allocated, ensuring that even in the worst-case scenario, this model can be executed as expected.

Note that multiple feature groups fail altogether, so if we want to cover all possible failure scenarios, this would imply that several fallback models should be trained considering all combinations of feature groups that can be unavailable. In a practical scenario the operational overhead of having these models may not pay off and a criteria to select the minimum number of fallback alternatives is needed. Consider you have a $N$ set of feature groups then the maximum number of fallback models is equal to the factorial $N!$. Note that the minimum number of fallback models to cover all feature group failure scenarios is equal to the feature group number, as you will always keep the fallback models that use a single feature group. For example, consider three feature groups, A, B and C. If we develop fallback models considering any of the feature groups can be unavailable, we have 6 fallback models, one using only A and B, other using B and C, other using A and C and other 3 models using only A, B and C respectively. Intuitively, and because the performance of these models can be evaluated beforehand, it is encouraged to only keep a small set of models that demonstrate higher performance, avoiding additional overhead. If we assert the performance of all 6 models and realize that the model using only C feature group has equivalent performance to the one using B and C, we should only keep the model using C feature group. This process is similar to group-wise feature selection~\cite{Kuzudisli23} and requires the Machine Learning practitioner to specify a tolerance threshold to tell whether two models have equivalent performance.

When an incident is open and a feature group is flagged as unhealthy, it is trivial to route the requests to the proper fallback model, but this is not the case when the external data providers are healthy but the main model service is facing latency spikes or complete unavailability. In scenarios where the model service is under stress, for example seasonal or unexpected events that cause a sudden increase in the request volume, the response time variance may increase and take longer to respond, potentially causing timeouts. This scenario is what we call latency spikes, when for a small number of requests, the model service may take way longer to respond. If the response time fails to meet the client SLA, the client will timeout and stop waiting for the model service to respond. Since the main model can still respond in an acceptable time fashion for most requests, we shouldn’t route all the traffic to the fallback models, instead, we can apply a retry policy to call the main model again. Note there is a non-trivial trade off between the timeout threshold and the maximum number of retries. The higher the timeout threshold, which is how long you wait for the model service to respond, the less likely it is to timeout, however, if it does, you will have less remaining time to retry. To illustrate, consider a total response time SLA is 300ms. If you specify 100ms as a timeout threshold, you can only retry 2 times at most, because if the model timeouts all the 3 attempts, the sum will be your SLA. As an alternative, if the model timeouts all retries, you can use the best fallback alternative instead. Once all models are independent of each other, it is possible to call them in parallel with the main model. Thus, if the main model fails to respond, you already have the fallback scores available and avoid adding up to the total response time. On the other hand, calling all models in parallel can be a waste of infrastructure resources as you are replicating the full workload to save response time for a few handful transactions. One approach to reduce the redundant resources is to selectively choose based on the request attributes which ones you desire to have this level of redundancy. For example, in the financial domain, you often find applications where the financial value of each model inference can drastically differ. The financial loss of letting a fraud or risky loan go through is directly proportional to the amount of the payment or credit operation. It is reasonable to desire higher robustness for model inferences associated with high financial value, thus, we can selectively opt for higher redundancy in such cases.

\section{Open Banking Payments \& Conclusion}

We have applied our proposed hierarchical fallback architecture to improve the robustness of Trustly’s Payment product. Trustly is a leader at Open Banking payments in North America. It leverages Open Banking data to conduct risk analysis for millions payments every month. Trustly connects to more than 99\% financial institutions in the US. Each one of them represents an external data provider key for machine learning inference. Financial institutions are diverse and, thus, some may be more reliable than others when it comes to service levels. In case of failure or financial institution unavailability, feature groups may be affected or become completely unavailable, thus, potentially hurting model performance. In order to protect the operation from potential external factors, we applied our fallback architecture. First, we mapped all feature groups to a potential failure scenario. Escorting every risk main model, we managed to produce fallback models, each one considering the absence of a given feature group. Additionally, to account for severe infrastructure outage scenarios where all models, including fallback, become unavailable, we also produced a Client-side fallback model, which is hardcoded in the application server. In the worst case scenario, where only the main application server is up, we would still have a risk evaluation being performed.

\begin{figure*}
    \centering
    \includegraphics[width=1.0\linewidth]{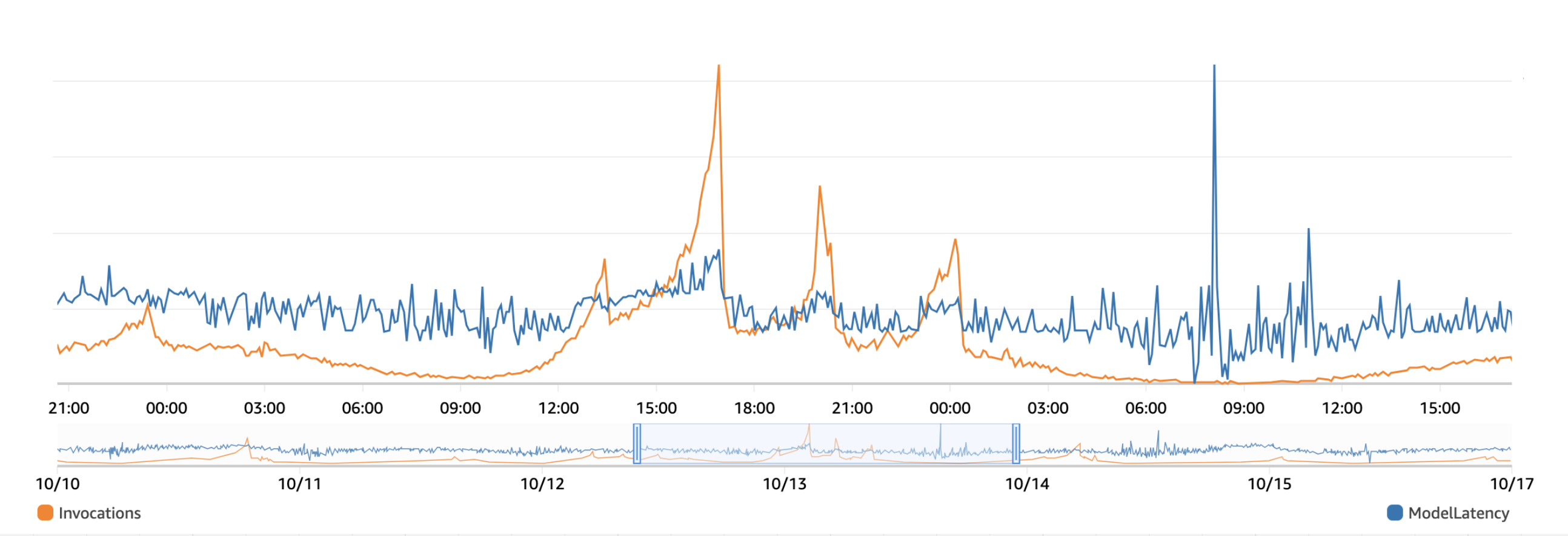}
    \caption{99th percentile model latency and number of requests on a 5 minutes aggregation window under the effect of a NFL game weekend.}
    \label{fig:latency-screenshot}
\end{figure*}

One shortcoming of our hierarchical fallback architecture is that the fallback models are expected to have lower performance compared to the main model. We observed that the biggest performance gap between the main model and a fallback model was about 9,09\% Precision-Recall Area Under the Curve (PR-AUC) and 58,06\% when compared to the hardcoded fallback model. Although the performance impact is significant when using hardcoded fallback model, it is important to note that less than 0,00001\% of the total volume is actually exposed to this scenario. 

Our most requested models operate and maintain robust Request Per Second (RPS) performance under regular business conditions, and can demonstrate significantly increased capacity during major events like the Super Bowl for our gaming customers. Figure~\ref{fig:latency-screenshot} illustrates the abrupt increase of request volume on a model service caused by a regular NFL game weekend. Seasonal events depend on the business vertical and are hard to accurately anticipate. Even on low request volume periods, a few handful requests may face timeouts and fail. Our hierarchical fallback structure contributes to mitigate potential risks and ensure robust ML operations even under extreme stress conditions.

Future work should focus on improving and evaluating retry policies and heuristics. It is not trivial to define timeout thresholds and maximum retry number for different stress scenarios. We consider that future research should provide evaluation mechanisms based on simulation to properly evaluate retry policies. Our hierarchical fallback architecture also should benefit from simulation based evaluation procedures. Finally, it is non trivial to select the the desired fallback models to balance the trade off between performance drop and failure scenario coverage. Future approaches should explore operational research techniques, such as maximum coverage~\cite{Nemhauser78}.

\begin{acks}
 The authors of this work would like to thank Trustly for all support.

\end{acks}

\bibliographystyle{ACM-Reference-Format}
\bibliography{sample-base}


\end{document}